\documentclass[letterpaper]{article} 
\usepackage[submission]{aaai24}  
\usepackage{times}  
\usepackage{helvet}  
\usepackage{courier}  
\usepackage[hyphens]{url}  
\usepackage{graphicx} 
\usepackage{natbib}  
\usepackage{caption} 
\usepackage{algorithm}
\usepackage{algorithmic}

\usepackage{newfloat}
\usepackage{listings}
\DeclareCaptionStyle{ruled}{labelfont=normalfont,labelsep=colon,strut=off} 
\floatstyle{ruled}
\newfloat{listing}{tb}{lst}{}
\floatname{listing}{Listing}
\usepackage{amssymb, amsmath, amsthm, amsfonts}
\usepackage[caption=false]{subfig}
\usepackage{enumerate}
\usepackage{booktabs}

\newtheorem{proposition}{Proposition}
\newtheorem{definition}{Definition}
\newtheorem{theorem}{Theorem}

\def\A{\mbox{\bf{A}}}
\def\E{\mathcal{E}}
\def\EE{\mbox{\bf{E}}}
\def\G{\mathcal{G}}
\def\HH{\mbox{\bf{H}}}
\def\I{\mbox{\bf{I}}}
\def\LL{\mathcal{L}}
\def\M{\mathcal{M}}
\def\N{\mathcal{N}}
\def\PP{\mathcal{P}}
\def\R{\mathcal{R}}
\def\SS{\mathcal{S}}
\def\V{\mathcal{V}}
\def\W{\mbox{\bf{W}}}
\def\WW{\mathcal{W}}
\def\X{\mbox{\bf{X}}}
\def\Z{\mbox{\bf{Z}}}
\def\dd{\boldsymbol{d}}
\def\h{\boldsymbol{h}}
\def\x{\boldsymbol{x}}
\def\z{\boldsymbol{z}}

\def\mmu{\boldsymbol{\mu}}
\def\nnu{\boldsymbol{\nu}}
\def\sig{\boldsymbol{\sigma}}
\def\ttau{\boldsymbol{\tau}}

\def\GNN{\text{GNN}}
\def\KL{\text{KL}}

\def\FNN{\text{FNN}}
\def\Nor{\text{Normal}}
\def\exp{\text{exp}}
\newsavebox\CBox

\title{Isomorphic-Consistent Variational Graph Auto-Encoders \\for Multi-Level Graph Representation Learning}
\author{
    Hanxuan Yang\textsuperscript{\rm 1,2}, Qingchao Kong\textsuperscript{\rm 2,1}, Wenji Mao\textsuperscript{\rm 2,1}
}
\affiliations{
    \textsuperscript{\rm 1}School of Artificial Intelligence, University of Chinese Academy of Sciences, Beijing 100049, China\\
    \textsuperscript{\rm 2}Institute of Automation, Chinese Academy of Sciences, Beijing 100190, China\\

    \{yanghanxuan2020, qingchao.kong, wenji.mao\}@ia.ac.cn
}

\begin{document}

\maketitle

\begin{abstract}
	Graph representation learning is a fundamental research theme and can be generalized to benefit multiple downstream tasks from the node and link levels to the higher graph level. In practice, it is desirable to develop task-agnostic general graph representation learning methods that are typically trained in an unsupervised manner. Related research reveals that the power of graph representation learning methods depends on whether they can differentiate distinct graph structures as different embeddings and map isomorphic graphs to consistent embeddings (i.e., the isomorphic consistency of graph models). However, for task-agnostic general graph representation learning, existing unsupervised graph models, represented by the variational graph auto-encoders (VGAEs), can only keep the isomorphic consistency within the subgraphs of 1-hop neighborhoods and thus usually manifest inferior performance on the more difficult higher-level tasks. To overcome the limitations of existing unsupervised methods, in this paper, we propose the Isomorphic-Consistent VGAE (IsoC-VGAE) for multi-level task-agnostic graph representation learning. We first devise a decoding scheme to provide a theoretical guarantee of keeping the isomorphic consistency under the settings of unsupervised learning. We then propose the Inverse Graph Neural Network (Inv-GNN) decoder as its intuitive realization, which trains the model via reconstructing the GNN node embeddings with multi-hop neighborhood information, so as to maintain the high-order isomorphic consistency within the VGAE framework. We conduct extensive experiments on the representative graph learning tasks at different levels, including node classification, link prediction and graph classification, and the results verify that our proposed model generally outperforms both the state-of-the-art unsupervised methods and representative supervised methods designed for individual tasks.
\end{abstract}

\section{Introduction}

Graph representation learning aims to map the topology structures and node attribute features of relational data into a low-dimensional embedding space. Based on different granularities of the embeddings (a.k.a. representations), the graph representation learning tasks have been extensively studied by the graph neural networks (GNNs) at node level \citep{kipf2017semi, velivckovic2018graph, zeng2020graphsaint}, link level  \citep{zhang2018link, zhang2021labeling} and graph level \citep{xu2019powerful}. However, these deep learning-based methods usually employ an end-to-end supervised training paradigm and can only learn graph representations for specific downstream tasks. In contrast, there are many situations where the task-agnostic representations are preferred and/or the label information is unavailable. Thus it is desirable to leverage the more general graph representation learning methods that are typically trained in an unsupervised manner.

The graph representation learning problem requires the model to learn embeddings with increasingly complex structural information. Related research has shown that the power of graph representation learning methods can be characterized as to which extent they can solve the \textit{graph isomorphism} problem \cite{grohe2020graph}, that is, whether they can differentiate distinct graph structures as different embeddings and map topologically identical (a.k.a., isomorphic) graphs into consistent embeddings, which we refer to as the \textit{isomorphic consistency} of graph models in this paper.

\begin{figure}[t]
	\begin{center}
		\includegraphics[width=\columnwidth]{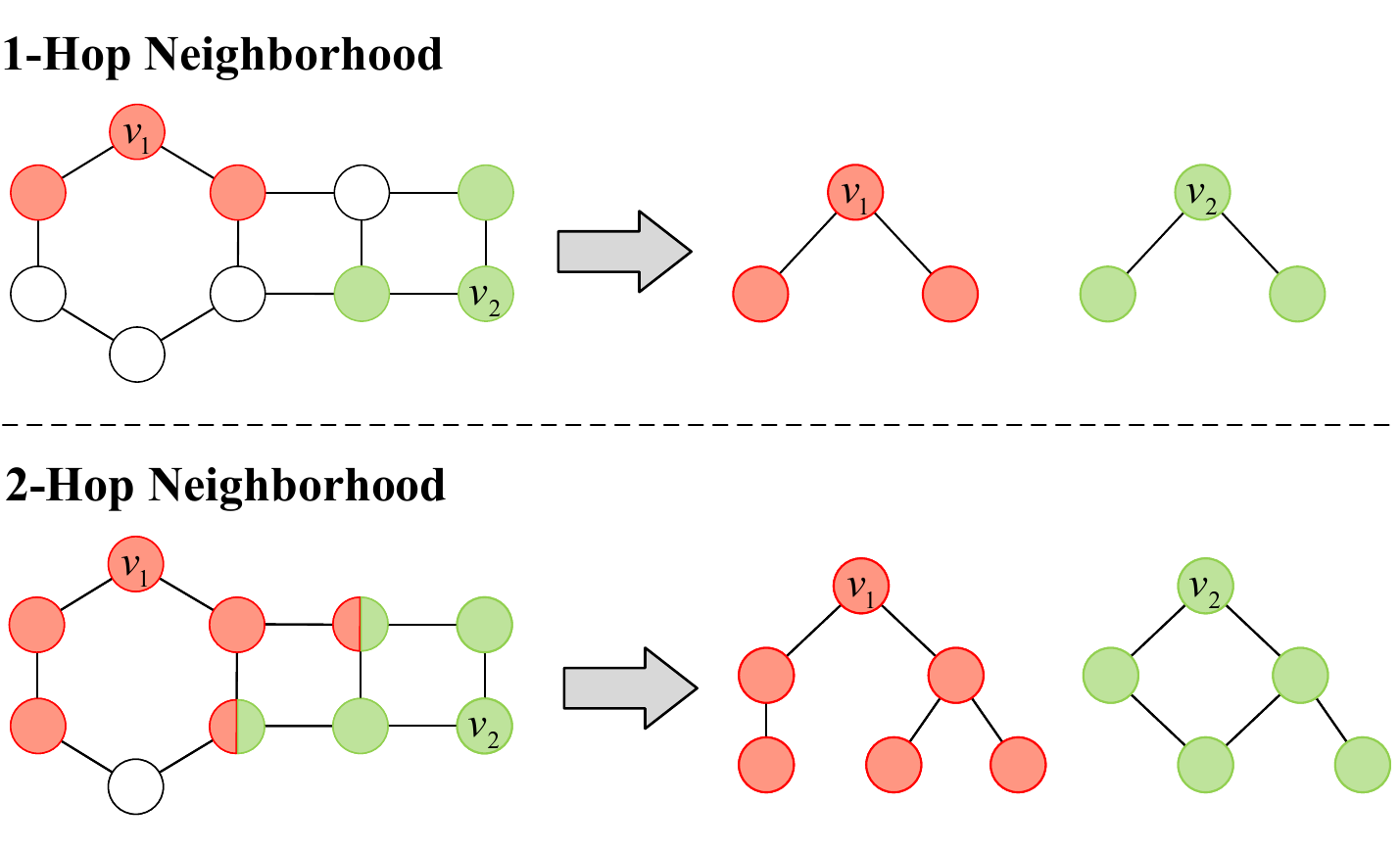}
		\caption{In this graph, nodes $v_1$ and $v_2$ have non-isomorphic global structures. To differentiate them as different embeddings, we should consider at least the 2-hop neighborhood information (bottom), since they share the same local structures within the 1-hop neighborhood (top).}
		\label{1-isomorphism}
	\end{center}
\end{figure}

Existing unsupervised graph representation learning methods are typically based on the deep generative models, represented by the variational graph auto-encoders (VGAEs) \citep{kipf2016variational}. These methods leverage a GNN-based encoder to aggregate the neighbor features as node embeddings. These embeddings can be applied to multiple downstream tasks \citep{bojchevski2018netgan, grover2019graphite, li2023maskgae}, as they are optimized in an unsupervised manner by reconstructing the edges of an adjacency matrix in the decoder, rather than using any task-specific label. However, since the graph edges only contain the 1-hop neighborhood information of each node, the embeddings learned by these VGAE-based methods can only keep the low-order isomorphic consistency, which is insufficient to learn distinguishable node embeddings for the more difficult higher-level tasks. For example, in Fig.~\ref{1-isomorphism}, nodes $v_1$ and $v_2$ should have different embeddings as they are at asymmetric positions in the global graph. However, if only 1-hop neighborhood subgraphs are considered, nodes $v_1$ and $v_2$ will have the same local structural information, thus they may learn consistent embeddings by mistake. Therefore, the capacity of VGAEs is hampered due to the lack of high-order isomorphic consistency, especially for graph-level tasks such as graph classification.

Previous research has developed several GNNs \cite{kipf2017semi, hamilton2017inductive, xu2019powerful, zhang2021labeling} that are isomorphic-consistent to different extents within the sense of the 1-dimensional Weisfeiler-Lehman test \cite{leman1968reduction}. These methods are capable of aggregating multi-hop neighborhood information as node embeddings according to graph topology in supervised learning. Intuitively, this inspires us to modify the decoder of VGAEs by reconstructing the GNN embeddings, instead of the adjacency matrix, so as to preserve the isomorphic consistency property of GNNs. To this end, we need to solve the challenging issues of identifying the theoretical criterion to ensure high-order isomorphic consistency of the decoder and developing computational means to maintain this property by node embedding reconstruction under unsupervised learning settings.

To address the above issues and overcome the limitations of VGAEs, in this paper, we propose the Isomorphic-Consistent VGAE (IsoC-VGAE) for multi-level graph representation learning. We first design a decoding scheme to provide a theoretical basis for keeping the high-order isomorphic consistency of the GNN-based encoder under unsupervised settings. We then develop a novel Inverse GNN (Inv-GNN) decoder as an instantiation of our proposed decoding scheme, which does the opposite work of the GNN-based encoder, that is, to reconstruct the self-embedding and neighborhood distribution of each node using the aggregated GNN embeddings. In this way, the high-order isomorphic consistency of GNNs can be properly preserved in the decoder by learning the multi-hop neighborhood information from the embeddings of the encoder. We conduct extensive experiments on the representative graph learning tasks at different levels, including node classification, link prediction and graph classification, and the results verify the effectiveness of our proposed model.

The main contributions of this work are as follows:
\begin{itemize}
	\item To tackle multi-level task-agnostic graph representation learning, we first propose the decoding scheme within the VGAE framework to theoretically ensure high-order isomorphic consistency in unsupervised learning.
	\item We develop a novel Inv-GNN decoder as a realization of the proposed decoding scheme, which trains the model via reconstructing the self-embeddings and neighborhood distributions learned by the GNN-based encoder.
	\item Experimental results on representative tasks at node, link and graph levels show that our model generally outperforms the state-of-the-art unsupervised and representative supervised graph representation learning methods.
\end{itemize}

\section{Preliminaries}

In this section, we briefly review two important graph representation learning methods that are used in our proposed model, namely the GNNs and VGAEs.

\subsection{Graph Neural Networks}

GNNs are deep learning-based methods for graph representation learning. They perform neighbor aggregation to learn node embeddings based on the graph topology and are trained in an end-to-end manner. Specifically, the node embeddings of the $l$-th GNN layer $\h_i^{(l)}$ are obtained as a weighted summation of the neighbor embeddings, i.e.,
\begin{align}
	\h_i^{(l+1)}=f(\W^{(l)}(\h_i^{(l)}+\sum_{j\in\N_i}\omega_{ij}^{(l)}\h_j^{(l)})),\label{gnn}
\end{align}
where $f(\cdot)$ is a nonlinear transformation function, $\N_i$ denotes the neighbor set of node $i$, and $\W^{(l)}$ is a trainable weight matrix. The embedding of each neighbor is weighted by $\omega_{ij}$, which is employed as, for example, the normalized graph Laplacian in the graph convolutional networks (GCNs) \citep{kipf2017semi}, and the masked attention weight in the graph attention networks (GATs) \citep{velivckovic2018graph}.

\subsection{Variational Graph Auto-Encoders}

VGAEs are among the most well-known and powerful unsupervised graph representation learning methods. They usually consist of a GNN-based encoder to learn node embeddings and a reconstruction decoder to train the model without any task-specific label. The vanilla VGAE \citep{kipf2016variational} generates node embeddings from Normal distributions and reconstructs the adjacency matrix $\A$ via the inner product, i.e.,
\begin{align}
	\hat{\A}_{ij}=\z_i^\top\z_j,\quad\z_i\sim\Nor(\mmu_i,\I\sig_i^2),
\end{align}
where $\hat{\A}=(\hat{\A}_{ij})$ denotes the reconstructed adjacency matrix, $\I$ is an identity matrix, $\mmu_i$ and $\sig_i$ are parameterized by a GCN encoder. For optimization, VGAEs minimize the negative evidence lower bound (ELBO), which can be formulated as the combination of a reconstruction loss and a Kullback-Leibler (KL) divergence, i.e.,
\begin{align}
	\LL_{vgae}=\LL_{rec}(\A,\hat{\A})+\sum_{i}\KL[q(\z_i\vert\h_i)\Vert p(\z_i)],
\end{align}
where $\KL[\cdot\Vert\cdot]$ denotes the KL divergence, $p(\cdot)$ and $q(\cdot)$ are the prior and variational posterior distributions, respectively.

\section{Isomorphic-Consistent Decoding Scheme}

In this section, we first provide the formal definition of task-agnostic general graph representation learning, and then propose a decoding scheme within the VGAE framework to learn powerful and general graph representations for multi-level tasks.

Considering an $N$ node undirected graph $\G=(\V,\E)$ and, if available, an attribute feature matrix $\X=(\x_1,\dots,\x_N)^\top$, where $\V=\{v_1,\dots,v_N\}$ and $\E=\{e_1,\dots,e_N\}$ are the sets of nodes and edges, respectively, the \textit{general graph representation learning} is to train an \textbf{unsupervised} graph model $\M$ to map the graph structure into a low-dimensional embedding space without any task-specific label, i.e., $\M:(\G,\X)\rightarrow\Z$, where $\Z=(\z_1,\dots,\z_N)^\top$ is the node embedding matrix. In the rest of this section, we shall omit the attribute features $\X$ for simplification, since we mainly focus on learning the structural information of nodes.

To explore the expressive power of an unsupervised graph model for multi-level general graph representation learning, we first give the definition of graph isomorphism.

\begin{definition}\label{definition1}
	Given two graphs $\G=(\V,\E)$ and $\G'=(\V',\E')$ with $N$ nodes, $\Pi_{\V'}$ is the permutation set including all $m!$ possible index permutations of the nodes in $\V'$. Graphs $\G$ and $\G'$ are \textbf{isomorphic}, denoted as $\G\simeq\G'$, if $\exists\pi_{\V'}\in\Pi_{\V'}$ such that $\V=\pi(\V')$ and $\E=\pi(\E')$, where $\pi_{\V'}(\cdot)$ indicates to reorder the nodes in $\V'$.
\end{definition}

Intuitively, the most powerful graph models should ensure that all non-isomorphic graphs or subgraphs formed by node and edge subsets always have distinct embeddings, and isomorphic graphs or subgraphs always have consistent embeddings. Next, we introduce the concept of isomorphic consistency to evaluate the expressive power of a graph model.

\begin{definition}\label{definition2}
	For graphs $\G=(\V,\E)$, $\G'=(\V',\E')$ and their node embeddings $\Z$, $\Z'$ learned by model $\M$, $\SS\subseteq\V$, $\SS'\subseteq\V'$ are two node subsets with $m$ nodes, $\G_{\SS}^{(k)}=(\V_{\SS}^{(k)},\E_{\SS}^{(k)})$, ${\G'}_{\SS'}^{(k)}=({\V'}_{\SS'}^{(k)},{\E'}_{\SS'}^{(k)})$ are subgraphs of $\G$ and $\G'$ that consist of the $k$-hop neighbors of the nodes in $\SS$ and $\SS'$, $k\geq 1$, and $\Z_{\SS}$, $\Z'_{\SS'}$ are the embeddings of the node subsets. Model $\M$ is \textbf{ $\boldsymbol{k}$-order isomorphic-consistent} if $\forall\SS\subseteq\V$, $\SS'\subseteq\V'$, $\Z_{\SS}=\Z'_{\SS'}\Leftrightarrow\G_{\SS}^{(k)}\simeq{\G'}_{\SS'}^{(k)}$.
\end{definition}

The node subsets $\SS$ and $\SS'$ can vary according to the objective task levels. For node-level tasks, $m=1$, for link-level tasks, $m=2$, and for graph-level tasks, $m=N$. The VGAEs can only ensure the 1-order isomorphic consistency since they learn node embeddings by reconstructing the edges of an adjacency matrix. Specifically, given $\Z^*$ as the optimal embeddings learned by VGAEs that can perfectly reconstruct all edges, $\Z^*$ can only guarantee the distinguishability of node subsets with different structural information in their 1-hop neighborhood subgraphs, since the graph edges merely contain the 1-hop neighborhood information of each node.

In addition, for graph-level tasks, the permutations of node indices should also be concerned since the node sets $\SS$ and $\SS'$ come from different graphs. Definition~\ref{definition2} implies that the isomorphic-consistent models should learn node embeddings that contain no node permutation information but only the topological structure of graphs. In other words, the isomorphic-consistent models should be able to map two isomorphic graphs with different node permutations into consistent embeddings. Formally, we give the definition of permutation invariance for unsupervised graph models as follows.

\begin{definition}\label{definition3}
	Letting $\Z$ be the node embeddings of graph $\G=(\V,\E)$ learned by model $\M$, $\M_{enc}$, $\M_{dec}$ and $\LL_{rec}$ be the encoder, decoder and reconstruction loss function of $\M$, respectively, model $\M$ is \textbf{permutation-invariant} to node orders if $\forall\pi_{\V}\in\Pi_{\V}$, $\M_{enc}(\G)=\M_{enc}(\pi_{\V}(\G))$, $\LL_{rec}(\M_{dec}(\Z))=\LL_{rec}(\M_{dec}(\pi_{\V}(\Z)))$.
\end{definition}

Most VGAE-based methods employ GNNs as the encoder $\M_{enc}$, which keep the permutation invariance in nature by aggregating node neighbors according to graph topology \citep{yang2019conditional}. However, the decoder $\M_{dec}$ of these methods are typically not permutation-invariant since they use the reconstruction error between the input adjacency matrix $\A$ and reconstructed adjacency matrix $\A'$ as the loss $\LL_{rec}$ for training, while $\exists\pi_{\V}\in\Pi_{\V}$, $\LL_{rec}(\A,\A')\neq\LL_{rec}(\A,\pi_{\V}(\A'))$. The lack of permutation invariance would waste the capacity of VGAEs in capturing the meaningless node permutations, instead of the underlying graph structure \citep{yang2019conditional}.

To leverage the VGAE framework for multi-level general graph representation learning, the main challenge is how to keep the high-order isomorphic consistency as well as the permutation invariance of the embeddings learned by the GNN-based encoder, while training the model under the settings of unsupervised learning in the decoder. The following proposition provides a decoding scheme for unsupervised graph representation learning methods to learn general embeddings that are theoretically as powerful as those learned by GNNs in terms of keeping the isomorphic consistency.

\begin{proposition}\label{proposition1}
	Given a graph $\G$ with attribute features $\X$, an unsupervised representation learning model $\M:(\G,\X)\rightarrow\Z$ and an $L$-layer $\GNN:(\G,\X)\rightarrow\HH^{(L)}$, model $\M$ is $L$-order isomorphic-consistent if the following conditions hold:
	\begin{enumerate}[a)]
		\item The GNN is $L$-order isomorphic-consistent.
		\item $\HH^{(L)}=f(\Z)$, where $f(\cdot)$ is an injective function.
	\end{enumerate}
\end{proposition}

\paragraph{Proof.} We start the proof with a simple case of the proposition, where the GNN only has one layer, i.e., $L=1$. With condition a), we have, $\forall\SS\subseteq\V$, $\SS'\subseteq\V'$,
\begin{align}
	\Z_{\SS}=\Z'_{\SS'}\Leftrightarrow&\HH_{\SS}^{(1)}=\HH_{\SS'}^{(1)}\nonumber\\
	\Leftrightarrow&\GNN_{\SS}(\G_{\SS_1},\HH^{(0)}_{\SS_1})=\GNN_{\SS'}(\G'_{\SS'_1},{\HH'}^{(0)}_{\SS'_1}),
\end{align}\label{simple_case}
where the subscript of the model $\GNN_{\SS}(\cdot)$ indicates the output embeddings of the nodes in $\SS$. Then, because the GNN is isomorphic-consistent (condition b)), we have
\begin{align}
	\Z_{\SS}=\Z'_{\SS'}\Leftrightarrow\G_{\SS_1}\simeq\G'_{\SS'_1}.
\end{align}

Now we consider the multi-layer case. With condition a), we have
\begin{align}
	\Z_{\SS}=\Z'_{\SS'}\Leftrightarrow\HH_{\SS}^{(L)}={\HH'}_{\SS'}^{(L)}.\label{multi_case}
\end{align}
Letting $\GNN_{\SS}^{(l)}(\cdot)$, $l=1,\dots,L$ denote the $l$-th layer of the GNN model, $\HH_{\SS}^{(L)}$ can be decomposed as
\begin{align}
	\HH_{\SS}^{(L)}=&\GNN_{\SS}^{(L)}(\G_{\SS_1},\HH_{\SS_1}^{(L-1)})\nonumber\\
	=&\GNN_{\SS}^{(L)}(\G_{\SS_1},\GNN_{\SS_1}^{(L-1)}(\G_{\SS_2}, \HH_{\SS_2}^{(L-2)}))\nonumber\\
	=&\GNN_{\SS}^{(L)}(\G_{\SS_1},\GNN_{\SS_1}^{(L-1)}(\nonumber\\
	&\cdots\GNN_{\SS_{L-1}}^{(1)}(\G_{\SS_L},\HH^{(0)}_{\SS_L})\cdots))\nonumber\\
	=&\GNN_{\SS}(\G_{\SS_L},\HH^{(0)}_{\SS_L}).
\end{align}

Similarly, 
\begin{align}
	{\HH'}_{\SS'}^{(L)}=&\GNN_{\SS'}^{(L)}(\G'_{\SS'_1},\GNN_{\SS'_1}^{(L-1)}(\nonumber\\
	&\cdots\GNN_{\SS'_{L-1}}^{(1)}(\G'_{\SS'_L},{\HH'}^{(0)}_{\SS'_L})\cdots))\nonumber\\
	=&\GNN_{\SS'}(\G'_{\SS'_L},{\HH'}^{(0)}_{\SS'_L}).
\end{align}
Therefore, for an $L$-layer GNN, the receptive fields of the node sets $\SS$ and $\SS'$ are the subgraphs $\G_{\SS_L}$ and $\G'_{\SS'_L}$, which include the $L$-hop neighborhoods of all nodes in $\SS$ and $\SS'$, respectively. Further, with condition b), we have
\begin{align}
	\GNN_{\SS}(\G_{\SS_L},\HH^{(0)}_{\SS_L})=\GNN_{\SS'}(\G'_{\SS'_L},{\HH'}^{(0)}_{\SS'_L})
	\Leftrightarrow\G_{\SS_L}\simeq\G'_{\SS'_L}.\label{multi_case2}
\end{align}
By combining Eq.~(\ref{multi_case})-(\ref{multi_case2}), we get the desired result.

The above decoding scheme enables us to build a VGAE-based model where the decoder can maintain the high-order isomorphic consistency (and so the permutation invariance) of the GNN encoder, as long as the GNN for the encoder can the embeddings $\Z$ learned by the decoder can conform to conditions a) and b) in Proposition~\ref{proposition1}, respectively.

For condition a), several GNNs have been proposed that are isomorphic-consistent to different extents, e.g., SEAL \citep{zhang2018link} and the graph isomorphism network (GIN) \citep{xu2019powerful}. To build decoders that can ensure condition b), one may use a more strict case where the injective function is constrained as the identity map. In the next section, we shall propose an instantiation of the decoders that can conform to condition b) by reconstructing the self-embeddings as well as the neighborhood distributions of the embeddings learned by the GNN encoder. Note that it is not the only choice based on the decoding scheme in Proposition~\ref{proposition1}, yet a simple and intuitive realization of our concern.

\section{Proposed Method}

\begin{figure}[t]
	\begin{center}
		\includegraphics[width=0.85\columnwidth]{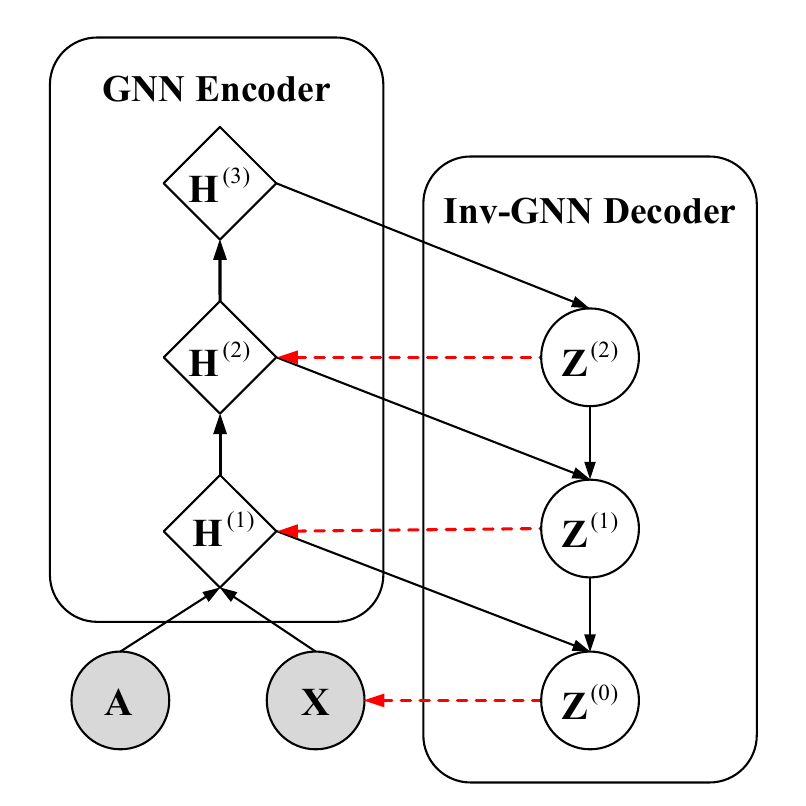}\par
		\caption{An illustration of the proposed IsoC-VGAE framework with three GNN layers. The black arrows indicate the feedforward propagation passes to learn node embeddings. The red dashed arrows indicate the reconstruction passes to train the model.}
		\label{G3M}
	\end{center}
\end{figure}

To learn general graph representations for multi-level tasks, we propose a generative graph model within the VGAE framework, i.e., the Isomorphic-Consistent VGAE (IsoC-VGAE), illustrated in Fig.~\ref{G3M}. Similar to vanilla VGAEs, the encoder employs GNNs to aggregate the multi-hop neighborhood information as node embeddings. Below we shall mainly introduce the decoder of our model.

\subsection{Inverse GNN Decoder}

\begin{figure}[t]
	\begin{center}
		\includegraphics[width=\columnwidth]{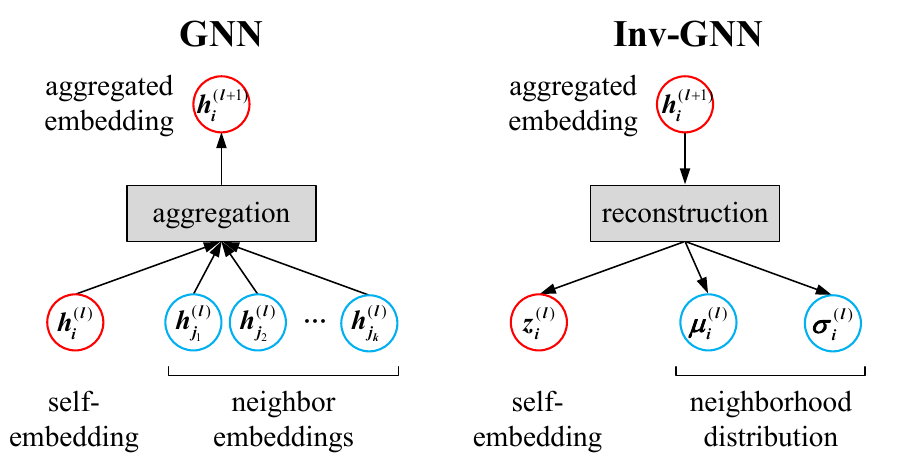}\par
		\caption{The message-passing mechanisms of a GNN (left) and the proposed Inv-GNN (right). A GNN layer aggregates the self-embedding and the $k$ neighbor embeddings into the next layer. An Inv-GNN layer uses the aggregated GNN embedding to reconstruct its self-embedding and neighborhood distribution of the previous layer.}
		\label{Inv-GNN}
	\end{center}
\end{figure}

Inspired by GNN layers which collect the self- and neighbor embeddings of each node as the aggregated embedding based on graph topology, we propose a novel inverse GNN (Inv-GNN) decoder to train our model using the opposite message-passing mechanism of GNNs, that is, reconstructing the self-embeddings as well as neighborhood distributions using the aggregated node embeddings, as presented in Fig.~\ref{Inv-GNN}.

\paragraph{Self-Embedding Reconstruction} Following the vanilla VGAEs, we leverage the reparameterization trick \citep{kingma2014auto} and generate the self-embeddings from Normal distributions parameterized by neural networks. Formally, for $l=0,\dots,L-1$, where $L$ is the number of layers in the GNN encoder, the reconstructed self-embeddings $\z_i^{(l)}\in\R^{C_l}$ of the $l$-th layer are obtained as
\begin{align}\label{self}
	\z_i^{(l)}\sim\Nor(\tilde{\mmu}^{(l)}_i+\mmu_i^{(l)},\I_{C_l}{\sig_i^{(l)}}^2),
\end{align}
where the variational posterior parameters are obtained using feed-forward neural networks (FNNs), i.e.,
\begin{align}
	\mmu_i^{(l)}=&\FNN_\mu(\h_i^{(l+1)}),\label{mean}\\
	\sig_i^{(l)}=&\exp(\FNN_\sigma(\h_i^{(l+1)})),\label{std}
\end{align}
$\tilde{\mmu}^{(l)}_i$ denotes the prior mean parameter set as $\FNN_z(\z_i^{(l+1)})$ for $l=0,\dots,L-2$ and zero for $l=L-1$, and $\I_{C_l}$ denotes the $C_l$-dimensional identity matrix.

Then, the self-embedding reconstruction loss is defined as
\begin{align}\label{loss_self}
	\LL_{self}=\sum_{l=0}^{L-1}\sum_{i=1}^{n}\Vert\h_i^{(l)}-\z_i^{(l)}\Vert_2^2,
\end{align}
where $\Vert\cdot\Vert_2$ denotes the $L^2$ norm, $\h_i^{(0)}$ is $\x_i$ if the node attribute features are available and a column vector of an identity matrix otherwise.

\paragraph{Neighborhood Reconstruction} The reconstruction of node neighbors is much more challenging, because the aggregated embeddings will inevitably lose some information from the neighbor embeddings during the pooling process, and the numbers of neighbors for different nodes may vary in a wide range. To this end, instead of reconstructing the specific values of the neighbor embeddings, our model attempts to learn the empirical distribution of each node neighborhood. Specifically, regarding each neighbor as an i.i.d. sample of the neighborhood distribution, the reconstruction of node neighbors can be transformed into inferring the parameters of the distribution as well as the number of samples (a.k.a., node degrees).

In our model, we learn the neighborhood distribution using the KL divergence between the variational posteriors of the reconstructed embeddings $q(\z_i^{(l)}\vert\h_i^{(l+1)})$ and the posteriors of the GNN embeddings $p(\h_i^{(l)}\vert\X)$, and the reconstruction loss of neighborhood distribution is defined as
\begin{align}\label{loss_neighbor}
	\LL_{nei}=\sum_{l=0}^{L-1}\sum_{i=1}^{n}\bigg(\KL[q(\z_i^{(l)}\vert\h_i^{(l+1)})\Vert p(\h_i^{(l)}\vert\X)]\bigg).
\end{align}
Here the KL divergence is only determined by the parameters of distributions but not related to the specific values of embedding samples. Thus, the proposed neighbor reconstruction strategy can avoid matching the permutation of the reconstructed neighbor embeddings to that of $\h_j^{(l)}$, which is not only computationally intractable but also non-permutation-invariant to node orders.

The posterior distribution of GNN embeddings $p(\h_i^{(l)}\vert\X)$ is assumed as a Normal distribution, of which the mean and variance parameters can be calculated as moments of the GNN embeddings within a neighborhood (including the self-embedding). In practice, we use a degenerate variance (i.e., the identity matrix) for the GNN embedding posterior, since We empirically find that such slight simplification of the neighborhood reconstruction hardly has any impact on model performance, but can bring a significant advantage on time efficiency.

To further learn the neighbor numbers of different nodes, our model also reconstructs the node degrees $d_i$, i.e.,
\begin{align}\label{degree}
	\hat{d}_i^{(l)}=f_+(\FNN_d(\h_i^{(l+1)})),
\end{align}
where $f_+(\cdot)$ denotes a non-negative transformation function (e.g., ReLU). The neighbor reconstruction loss is then defined as
\begin{align}\label{loss_degree}
	\LL_{deg}=\sum_{l=0}^{L-1}\sum_{i=1}^{n}\bigg(\Vert d_i-\hat{d}_i^{(l)}\Vert_2^2\bigg).
\end{align}

Last, the full loss function of our proposed model is defined as a combination of the self and neighbor reconstruction loss, i.e.,
\begin{align}\label{loss}
	\LL=\LL_{self}+\lambda_{nei}\LL_{nei}+\lambda_{deg}\LL_{deg},
\end{align}
where $\lambda_{nei}$ and $\lambda_{deg}$ are tuning hyperparameters to be specified. The pseudo-code for training our model is given in Algorithm~\ref{algorithm}.

\begin{algorithm}[tb]
	\caption{Training IsoC-VGAE}
	\label{algorithm}
	\textbf{Input}: Undirected graph $\G$ with $N$ nodes; node feature $\X$\\
	\textbf{Parameter}: Layer number $L$; neighbor tuning $\lambda_{nei}$; degree tuning $\lambda_{deg}$; learning rate $\kappa$; other hyperparameters\\
	\textbf{Output}: Node Embeddings $\HH^{(1:L)}$
	\begin{algorithmic}[1] 
		\STATE \textbf{Initialize} weight parameters $\W$
		\STATE $\HH^{(0)}=\X$
		\STATE $\tilde{\mmu}^{(L-1)}=\boldsymbol{0}$
		\WHILE{not convergence}
		\FOR{$l=0,\dots,L-1$}
		\STATE $\HH^{(l+1)}=\textsc{GNN}^{(l)}(\G,\HH^{(l)})$
		\ENDFOR
		\FOR{$l=L-1,\dots,0$}
		\STATE $\mmu^{(l)}=\textsc{FNN}_{\mu}^{(l)}(\HH^{(l+1)})$
		\STATE $\sig^{(l)}=\exp(\textsc{FNN}_{\sigma}^{(l)}(\HH^{(l+1)}))$
		\STATE $\hat{\dd}^{(l)}=f_{+}(\textsc{FNN}_{d}^{(l)}(\HH^{(l+1)}))$
		\STATE $\Z^{(l)}=\textsc{NormalSampling}(\tilde{\mmu}^{(l)}+\mmu^{(l)},\sig^{(l)})$
		\STATE $\tilde{\mmu}^{(l-1)}=\FNN_z(\Z^{(l)})$
		\ENDFOR
		\STATE $\LL_{self}=\sum_{l=0}^{L-1}\Vert\HH^{(l)}-\Z^{(l)}\Vert_2^2$
		\STATE $\LL_{nei}=\sum_{l=0}^{L-1}\KL[q(\Z^{(l)}\vert\HH^{(l+1)})\Vert p(\HH^{(l)}\vert\X)]$
		\STATE $\LL_{deg}=\sum_{l=0}^{L-1}\Vert\dd-\hat{\dd}^{(l)}\Vert_2^2$
		\STATE $\LL=\LL_{self}+\lambda_{nei}\LL_{nei}+\lambda_{deg}\LL_{deg}$
		\STATE $\W\gets\W-\kappa\nabla_{W}\mathcal{L}$
		\ENDWHILE
	\end{algorithmic}
\end{algorithm}

\subsection{Decoder Capacity Analysis}

Our proposed Inv-GNN decoder is constructed based on the isomorphic-consistent decoding scheme in Proposition~\ref{proposition1}. To conform to condition a), the Inv-GNN decoder employs FNNs to reconstruct the self-embedding and neighborhood distribution of each GNN embedding. Specifically, for the self-embedding reconstruction, FNNs have been widely adopted in learning such real-valued embeddings. As for the neighborhood reconstruction, we have the following theorem.

\begin{theorem}\label{theorem}
	Let $\Phi_0$ and $\Phi_1$ denote two $D$-dimensional diagonal multivariate Normal distributions, where the support of $\Phi_0$ is bounded. Given an arbitrarily small positive approximation error $\epsilon\rightarrow0^+$, there exists a fully connected and feed-forward neural network FNN $u(\cdot):\R^D\rightarrow\R^D$ with sufficiently large depth and width (depending on $D$, $\epsilon$ and $\EE_{\x\sim\Phi_0}\Vert\x\Vert^3$) such that $\KL[{\nabla_u}_\#\Phi_1\Vert\Phi_0]<\epsilon$, where ${\nabla_u}_\#$ indicates transforming a distribution using the gradient of $u(\cdot)$.
\end{theorem}

\paragraph{Proof.} The conclusion of Theorem~\ref{theorem} is derived from Theorem~3.1 in \cite{tang2022graph}, which we restate below.

\noindent\textit{For any $\epsilon>0$, if the support of distribution $\PP$ lies in a bounded space of $\R^D$, there exists an FNN $u(\cdot):\R^D\rightarrow\R^D$ (and thus its gradient $\nabla_u(\cdot):\R^D\rightarrow\R^D$) with large enough width and depth (depending on $d$, $\epsilon$ and $\EE_{\x\sim\PP}\Vert\x\Vert^3$) such that $\WW_2(\PP,{\nabla_u}_\#\Phi)^2<\epsilon$, where $\WW_2(\cdot,\cdot)^2$ is the 2-Wasserstein distance between distributions and $\Phi$ is a $d$-dimensional non-degenerate Normal distribution.}

Based on the above theorem, we only need to show the connection between the 2-Wasserstein distance and the KL divergence for diagonal multivariate Normal distributions. Let ${\nabla_u}_\#\Phi_1=\Nor(\mmu,\I_d\sig^2)$ and $\Phi_0=\Nor(\nnu,\I_d\ttau^2)$ be the source and transformed target distributions, respectively. The 2-Wasserstein distance between $\Phi_0$ and ${\nabla_u}_\#\Phi_1$ can be formulated as
\begin{align}
	\WW_2(\Phi_0,{\nabla_u}_\#\Phi_1)^2&=\Vert\nnu-\mmu\Vert_2^2+\Vert\ttau-\sig\Vert_2^2\nonumber\\
	&=\sum_{d=1}^D\bigg((\nu_d-\mu_d)^2+(\tau_d-\sigma_d)^2\bigg),
\end{align}
where $\mmu=(\mu_1,\dots,\mu_D)^\top$, $\sig=(\sigma_1,\dots,\sigma_D)^\top$, $\nnu=(\nu_1,\dots,\nu_D)^\top$, $\ttau=(\tau_1,\dots,\tau_D)^\top$. When $\epsilon\rightarrow0^+$, we have
\begin{align}
	&0\leq\WW_2({\nabla_u}_\#\Phi_1,\Phi_0)^2<0^+\nonumber\\
	\Rightarrow&\WW_2(\Phi_0,{\nabla_u}_\#\Phi_1)^2=0\nonumber\\
	\Rightarrow&\mu_d=\nu_d,\sigma_d=\tau_d,d=1,\dots,D.\label{wasserstein}
\end{align}
Now we expand the KL divergence between $\Phi_0$ and ${\nabla_u}_\#\Phi_1$ as
\begin{align}
	\KL[{\nabla_u}_\#\Phi_1\Vert\Phi_0]=&\frac{1}{2}\sum_{d=1}^D\bigg(\frac{(\mu_d-\nu_d)^2}{\tau_d^2}+\frac{\sigma_d^2}{\tau_d^2}\nonumber\\
	&+2(\ln\tau_d-\ln\sigma_d)-1\bigg).\label{KL}
\end{align}
Combining Eq.~(\ref{wasserstein}) and (\ref{KL}), we have
\begin{align}
	\lim\limits_{\epsilon\rightarrow0^+}\KL[\Phi_0\Vert{\nabla_u}_\#\Phi_1]=0.
\end{align}
So we get the desired result.

Assuming the GNN embedding posteriors $p(\h_i^{(l)}\vert\X)$ as diagonal multivariate Normal distributions (which is a standard assumption of GNNs), the above conclusion theoretically guarantees the feasibility for reconstructing $p(\h_i^{(l)}\vert\X)$ by minimizing the KL divergence between $q(\z_i^{(l)}\vert\h_i^{(l+1)})$ and $p(\h_i^{(l)}\vert\X)$ in Eq.~(\ref{loss_neighbor}), where the parameters of $q(\z_i^{(l)}\vert\h_i^{(l+1)})$ are obtained by transforming the embedding samples from $p(\h_i^{(l+1)}\vert\X)$ using FNNs in Eq.~(\ref{mean}) and (\ref{std}).

In addition, the proposed Inv-GNN decoder can also maintain the permutation invariance of GNNs, since both the self-embedding and neighborhood reconstructions do not use the adjacency matrix in the loss function and are independent of the permutations of node indices. This property ensures that our model focuses on learning the embeddings that can best represent the graph structural information instead of the meaningless node permutations.

\section{Related Work}

In this section, we briefly review the literature related to the expressive power of GNNs and the representative work of unsupervised graph representation learning methods.

\subsection{GNN-Based Graph Representation Learning}

GNNs are the most popular graph models for supervised graph representation learning. Representative GNN-based methods include GCN \citep{kipf2017semi} and GraphSAGE \citep{hamilton2017inductive}, both of which can only perform well on node-level tasks such as node classification but are not isomorphic-consistent for learning the higher-level embeddings. SEAL \citep{zhang2018link} leverages the labeling trick to keep the isomorphic consistency for learning embeddings at the multi-node (e.g., link) level. \citet{xu2019powerful} proposes GIN for graph-level representation learning and first employs the 1-dimensional Weisfeiler-Lehman (1-WL) test \cite{leman1968reduction} to evaluate the isomorphic consistency of GNNs, which is soon widely adopted by follow-up work \cite{morris2019weisfeiler, chen2019equivalence, maron2019provably, sato2019approximation, li2020distance, sato2021random, zhang2021labeling, wang2022powerful}. In addition to the 1-WL test, there are also some work attempts to measure the expressive power of GNNs in other ways \citep{maron2019universality, keriven2019universal, loukas2020graph, garg2020generalization, chen2020graph, chen2021graph}. All of these studies are based on the GNN frameworks, which are typically designed for some individual downstream task and must be retrained (or even redesigned) when applied to a new task.

\subsection{Unsupervised Graph Representation Learning}

Early unsupervised graph representation learning methods mainly focus on link-level tasks. \citet{kipf2016variational} first proposes the VGAEs that reconstruct the adjacency matrix for link prediction, which has triggered plenty of successive work in this direction \citep{pan2018adversarially, bojchevski2018netgan, grover2019graphite, mehta2019stochastic, sarkar2020graph, hou2022graphmae, li2023maskgae}. Besides, GraphGAN \citep{wang2018graphgan, zhu2021adversarial} leverages the generative adversarial net (GAN) framework to learn general graph embeddings. However, all these methods can only preserve the 1-order isomorphic consistency and are not permutation-invariant, as they are trained by reconstructing the edges in an adjacency matrix which merely includes the local topological structure within 1-hop neighborhoods. In particular, \citet{tang2022graph} proposes the neighborhood Wasserstein reconstruction (NWR), which reconstructs the multi-hop GNN embeddings of node neighbors using the Wasserstein distance and thus can keep the higher-order topological structures. Nevertheless, this method is still not isomorphic-consistent since it is not permutation-invariant to the order of nodes, and must match the permutations of the GNN embeddings and reconstructed embeddings for training. There are also non-generative methods for unsupervised graph representation learning. For example, the deep graph infomax (DGI) \citep{velivckovic2019deep} and InfoGraph \citep{sun2020infograph} proposed for node- and graph-level tasks, respectively, employ the contrastive learning method and maximize the mutual information between embeddings, but are not general enough for multi-level graph representation learning, as they need to construct various types of negative samples for different tasks. To the best of our knowledge, there is still a lack of research on studying the isomorphic consistency of unsupervised graph representation learning methods.

\section{Experiments}

We evaluate the effectiveness of our proposed model on three representative graph representation learning tasks of different granularities, including node classification (node-level), link prediction (link-level) and graph classification (graph-level).

\subsection{Datasets}

The experiments are conducted on well-known benchmark datasets of each task. Specifically, for node classification and link prediction, we use the citation networks, i.e., Cora, CiteSeer and PubMed \cite{sen2008collective}, all of which are constructed using documents from different disciplines as nodes and the citation links between them as edges. Each citation network has a sparse binary feature matrix obtained using the bag-of-words features (for Cora and PubMed) or one-hot category (for CiteSeer) of each document. The node labels are employed using the disciplines the documents belong to. Following the standard supervised node classification settings, for each network, we use the default node splits for testing and validation, and all other nodes are used for training. As for link prediction, we randomly split the edges of each network as 85\% for training, 10\% for testing and 5\% for validation. Additionally, we also consider two large-scale datasets, i.e., Flickr \cite{zeng2020graphsaint} for node classification and ogbl-collab \cite{wang2020microsoft} for link prediction. Flickr is formed by images collected from websites, of which the edges denote whether two images share any common metadata. The node features and classes are employed using the descriptions and types of the images, respectively. The ogbl-collab is a large-scale collaboration network of the open graph benchmark (OGB) datasets for link prediction. The nodes represent authors and the edges denote collaborations between them. This dataset includes a node feature matrix obtained using word embeddings of the papers published by the authors. Both of the two large-scale datasets use the default data splits. The descriptive statistics of the datasets for node classification and link prediction are given in Table~\ref{datasets_link}.

For graph classification, we use two molecular datasets, i.e., MUTAG and PTC-MR \cite{yanardag2015deep}, and two social network datasets, i.e., IMDB-BINARY and COLLAB \cite{yanardag2015deep}. MUTAG and PTC-MR consist of chemical compounds, where the nodes represent atoms and edges represent chemical bonds. The molecules in MUTAG are labeled according to their mutagenic effect on a bacterium, and those in PTC-MR are labeled based on their carcinogenicity on male rats. The types of atoms in the molecules are employed as one-hot node features for both of the datasets, and the edge types are omitted in our experiments. IMDB-B and COLLAB are collaboration datasets, where each graph is an ego-network of a movie or a researcher, respectively. The nodes in each graph of IMDB-B represent actors/actresses and the edges represent whether they appear in the same movie. The graphs are labeled according to the genre of the movie. The nodes in each graph of COLLAB represent researchers and the edges represent the collaborations between them. The graphs are labeled based on the fields the researchers belong to. We randomly split the graphs of each dataset as 50\% for training, 30\% for testing and 20\% for validation. The descriptive statistics of the datasets for graph classification are given in Table~\ref{datasets_graph}.

\begin{table}[t]
	\centering
	\resizebox{\columnwidth}{!}{
		\begin{tabular}{lcccc}
			\toprule
			\rule{0pt}{10pt}  &\textbf{\# Nodes} &\textbf{\# Edges}    &\textbf{\# Features} &\textbf{\# Node Classes}\\
			\midrule
			\rule{0pt}{9pt}Cora       & 2,708 &     5,278& 1,433      &6\\[1pt]
			CiteSeer   & 3,312 &     4,552& 3,703      &7\\[1pt]
			PubMed     &19,717 &    44,324& 500        &3\\[1pt]
			Flickr     & 89,250&  899,756&  500        &7\\[1pt]
			ogbl-collab&235,868& 1,285,465& 128        &--\\
			\bottomrule
		\end{tabular}
        }
        \caption{Descriptive statistics of datasets for node classification and link prediction.}\label{datasets_link}
\end{table}

\begin{table}[t]
	\centering
	\resizebox{\columnwidth}{!}{
		\begin{tabular}{lcccc}
			\toprule
			\rule{0pt}{9pt}    &\textbf{\# Graphs}&\textbf{Avg. \# Nodes}&\textbf{\# Features} &\textbf{\# Graph Classes}\\
			\midrule
			\rule{0pt}{9pt}MUTAG      &188     &     17.9   &7&2\\[1pt]
			PTC-MR     &344     &     14.3   &18&2\\[1pt]
			IMDB-B     &1,000 &     19.8   &--&2\\
			COLLAB     &5,000 &     74.5   &--&3\\
			\bottomrule
		\end{tabular}
        }
        \caption{Descriptive statistics of datasets for graph classification.}\label{datasets_graph}
\end{table}

\subsection{Baselines}

For comparative methods, we consider the state-of-the-art unsupervised graph representation learning methods, including four VGAE-based methods, i.e., the vanilla (V)GAE \citep{kipf2016variational}, PIGAE \citep{winter2021permutation}, NWR \citep{tang2022graph} and the very recent MaskGAE \citep{li2023maskgae}, and two contrastive learning-based methods, i.e., DGI \citep{velivckovic2019deep} (for node classification and link prediction) and InfoGraph \citep{sun2020infograph} (for graph classification). In addition, we also consider the supervised multilayer perceptron (MLP) and several representative GNN-based methods, including GCN \citep{kipf2017semi} and GAT \cite{velivckovic2018graph} for node classification, SEAL \cite{zhang2018link} for link prediction, and GIN \citep{xu2019powerful} for graph classification. Note that GCN and GIN are also employed as the encoders of the VGAE-based methods for the node/link- and graph-level tasks, respectively.

\subsection{Experimental Settings}
We train all supervised comparative methods in an end-to-end manner for each task. As for the unsupervised methods, including our proposed IsoC-VGAE, we first train the models in an unsupervised manner and then feed the learned embeddings to MLPs for different supervised tasks. The link- and graph-level embeddings are obtained by computing the inner products and sums of the learned node embeddings, respectively. In particular, we perform link prediction in an end-to-end manner for the methods that are trained using adjacency matrices, including (V)GAE, PIGAE and MaskGAE, as they were originally designed to do. Since we mainly focus on evaluating the power of decoders, for all VGAE-based methods, including (V)GAE, NWR, MaskGAE and our proposed IsoC-VGAE, we use GCN as the encoder for node classification and link prediction, and GIN as the encoder for graph classification. 

For the hyperparameters of our model, we use a two-layer GNN encoder with the channel dimension of 512, and the same for the Inv-GNN decoder. The tuning parameters $\lambda_{nei}$ and $\lambda_{deg}$ are set as 0.1 and 1, respectively. The MLPs for supervised tasks are set as four layers with the channel dimension of 256.

\begin{table}[t]
	\centering
	\resizebox{\columnwidth}{!}{
		\begin{tabular}{lcccc}
			\toprule
			\rule{0pt}{10pt}&\textbf{Cora}&\textbf{CiteSeer}&\textbf{PubMed}&\textbf{Flickr}\\
			\midrule
			\rule{0pt}{9pt}MLP&70.8 $\pm$ 1.6&71.7 $\pm$ 0.9&86.8 $\pm$ 0.6&47.2 $\pm$ 0.1\\[1pt]
			GCN&\textbf{86.3 $\pm$ 0.3}&\underline{77.0 $\pm$ 0.3}&\underline{87.2 $\pm$ 0.5}&\underline{51.0 $\pm$ 0.3}\\[1pt]
			GAT&86.0 $\pm$ 0.4&76.8 $\pm$ 0.3&85.8 $\pm$ 0.1&50.6 $\pm$ 0.5\\
			\midrule
			\rule{0pt}{9pt}GAE&72.1 $\pm$ 2.5&57.1 $\pm$ 1.6&73.2 $\pm$ 0.9&44.6 $\pm$ 0.3\\[1pt]
			VGAE&72.9 $\pm$ 1.5&60.8 $\pm$ 1.9&81.3 $\pm$ 0.8&44.3 $\pm$ 0.2\\[1pt]
			DGI&83.6 $\pm$ 0.6&73.8 $\pm$ 1.0&86.9 $\pm$ 0.3&48.8 $\pm$ 0.4\\[1pt]
			PIGAE&80.2 $\pm$ 0.5&68.9 $\pm$ 0.6&81.4 $\pm$ 0.6&46.5 $\pm$ 0.3\\[1pt]
			NWR&83.6 $\pm$ 1.6&71.5 $\pm$ 2.4&83.4 $\pm$ 0.9&50.3 $\pm$ 0.1\\[1pt]
			MaskGAE&86.0 $\pm$ 0.7&76.5 $\pm$ 1.1&84.3 $\pm$ 0.9&49.4 $\pm$ 0.2\\
			\midrule
			\rule{0pt}{9pt}\textbf{IsoC-VGAE}&\textbf{86.3 $\pm$ 0.4}&\textbf{77.2 $\pm$ 0.3}&\textbf{87.7 $\pm$ 0.2}&\textbf{51.2 $\pm$ 0.1}\\
			\bottomrule
		\end{tabular}
	}
	\caption{Node classification accuracy (\%) for different methods. The best results are in bold and the second best ones are underlined.}\label{results_node}
\end{table}

\begin{table}[tb]
	\centering
	\resizebox{\columnwidth}{!}{
		\begin{tabular}{lcccc}
			\toprule
			\rule{0pt}{10pt}&\textbf{Cora}&\textbf{CiteSeer}&\textbf{PubMed}&\textbf{ogbl-collab}\\
			\midrule
			\rule{0pt}{9pt}MLP&80.7 $\pm$ 0.6&88.8 $\pm$ 0.6&91.6 $\pm$ 0.2&94.0 $\pm$ 0.1\\[1pt]
			SEAL&92.2 $\pm$ 1.1&93.4 $\pm$ 0.5&93.0 $\pm$ 1.0&\textbf{95.7 $\pm$ 0.1}\\
			\midrule
			\rule{0pt}{9pt}GAE&91.2 $\pm$ 0.7&90.5 $\pm$ 0.5&95.7 $\pm$ 0.2&94.1 $\pm$ 0.1\\[1pt]
			VGAE&91.4 $\pm$ 0.0&90.8 $\pm$ 0.0&94.4 $\pm$ 0.0&94.5 $\pm$ 0.1\\[1pt]
			DGI&91.4 $\pm$ 0.8&93.2 $\pm$ 0.8&96.9 $\pm$ 0.0&95.0 $\pm$ 0.1\\[1pt]
			PIGAE&89.8 $\pm$ 0.6&90.6 $\pm$ 0.9&94.1 $\pm$ 0.1&92.6 $\pm$ 0.2\\[1pt]
			NWR&91.7 $\pm$ 0.6&92.2 $\pm$ 0.8&98.0 $\pm$ 0.1&93.6 $\pm$ 0.6\\[1pt]
			MaskGAE&\textbf{94.9 $\pm$ 0.5}&\underline{95.2 $\pm$ 1.0}&\textbf{98.3 $\pm$ 0.0}&94.6 $\pm$ 0.4\\
			\midrule
			\rule{0pt}{9pt}\textbf{IsoC-VGAE}&\underline{93.3 $\pm$ 0.3}&\textbf{95.3 $\pm$ 0.4}&\underline{98.1 $\pm$ 0.1}&\underline{95.2 $\pm$ 0.0}\\
			\bottomrule
		\end{tabular}
        }
        \caption{Link prediction AUC (\%) for different methods. The best results are in bold and the second best ones are underlined.}\label{results_link}
\end{table}

\subsection{Results}

We employ the widely used classification accuracy as the evaluation metric for node and graph classification, and the area under the ROC curve (AUC) for link prediction. All results are reported as the means and standard deviations over 5 independent runs.

\paragraph{Node Classification} Table~\ref{results_node} demonstrates that our model can significantly outperform all unsupervised comparative methods. Note that the vanilla (V)GAE even shows inferior performance to the MLP (except for the Cora dataset), indicating that the node embeddings learned by reconstructing the adjacency matrix can perform worse than the raw node features. On the other hand, our model can not only significantly outperform the MLP, but also show comparable performance to the GNN-based supervised methods, which further verifies that the proposed Inv-GNN decoder can well keep the representation learning power of GNNs.

\begin{table}[t]
	\centering
	\resizebox{\columnwidth}{!}{
		\begin{tabular}{lcccc}
			\toprule
			\rule{0pt}{10pt}&\textbf{MUTAG}&\textbf{PTC-MR}&\textbf{IMDB-B}&\textbf{COLLAB}\\
			\midrule
			\rule{0pt}{9pt}MLP&88.6 $\pm$ 4.1&56.5 $\pm$ 2.0&52.2 $\pm$ 1.2&47.2 $\pm$ 0.1\\[1pt]
			GIN&89.0 $\pm$ 1.5&\underline{62.1 $\pm$ 5.3}&\underline{72.5 $\pm$ 1.2}&\underline{67.5 $\pm$ 1.9}\\
			\midrule
			\rule{0pt}{9pt}GAE&73.9 $\pm$ 5.3&52.4 $\pm$ 4.5&59.6 $\pm$ 2.9&59.7 $\pm$ 1.7\\[1pt]
			VGAE&71.1 $\pm$ 2.3&58.8 $\pm$ 3.8&53.8 $\pm$ 1.0&58.9 $\pm$ 3.4\\[1pt]
			InfoGraph&87.5 $\pm$ 1.3&58.3 $\pm$ 3.0&64.9 $\pm$ 1.0&65.9 $\pm$ 0.3\\[1pt]
			PIGAE&82.5 $\pm$ 2.5&56.6 $\pm$ 2.3&62.2 $\pm$ 1.1&64.1 $\pm$ 0.9\\[1pt]
			NWR&\underline{89.3 $\pm$ 1.3}&61.6 $\pm$ 1.7&72.4 $\pm$ 0.3&68.5 $\pm$ 0.1\\[1pt]
			MaskGAE&85.0 $\pm$ 2.7&59.0 $\pm$ 1.7&58.3 $\pm$ 5.1&57.8 $\pm$ 0.7\\
			\midrule
			\rule{0pt}{9pt}\textbf{IsoC-VGAE}&\textbf{89.6 $\pm$ 2.0}&\textbf{62.5 $\pm$ 1.1}&\textbf{73.2 $\pm$ 0.9}&\textbf{70.1 $\pm$ 0.8}\\[1pt]
			\bottomrule
		\end{tabular}
	}
	\caption{Graph classification accuracy (\%) for different methods. The best results are in bold and the second best ones are underlined.}\label{results_graph}
\end{table}

\paragraph{Link Prediction} The experimental results in Table~\ref{results_link} show that our model achieves at least comparable results to both the supervised and unsupervised methods for link prediction. It should be mentioned that most of the comparative methods here, including SEAL, (V)GAE, DGI, PIGAE and MaskGAE, should be especially powerful for this task as they use edges as training labels in the loss functions. The experimental results also verify that our proposed Inv-GNN decoder can generally reserve the power of VGAEs for the link-level task, and meanwhile have much better generalizability to other levels of graph representation learning tasks.

\paragraph{Graph Classification} From Table~\ref{results_graph}, it can be seen that our proposed IsoC-VGAE can significantly outperform all of the comparative unsupervised graph representation learning methods, most of which only perform comparably to the MLP baseline due to their poor generalizability to the graph-level tasks. In addition, our model also achieves comparable results to the supervised method GIN, which verifies the effectiveness of the proposed Inv-GNN decoder as well as the theoretical decoding scheme for preserving the high-order isomorphic consistency of the GIN encoder to learn graph-level representations.

\subsection{Ablation Study}

We further conduct a series of experiments to evaluate different reconstruction components of the proposed Inv-GNN decoder, including the neighborhood distribution and node degrees. The experimental results on node classification are presented in Table~\ref{ablation_node}-\ref{ablation_graph}, which demonstrate that the proposed neighbor reconstruction strategy via learning the neighborhood distributions and node degrees is effective for improving the performance on downstream tasks. 

\begin{figure*}[htbp]
	\centering
	\subfloat[Node Classification]{
		\includegraphics[width=0.3\textwidth]{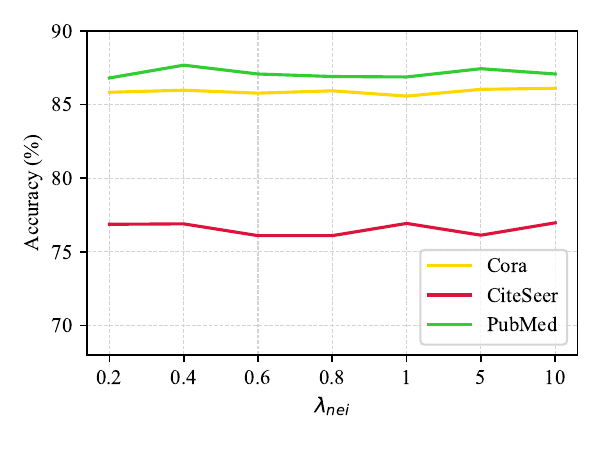}
	}
	\subfloat[Link Prediction]{
		\includegraphics[width=0.3\textwidth]{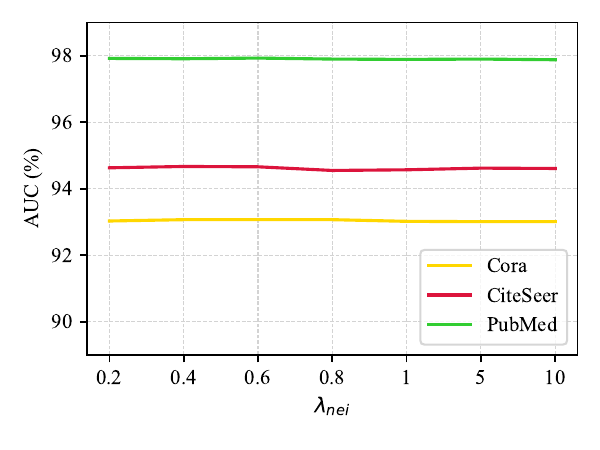}
	}
	\subfloat[Graph Classification]{
		\includegraphics[width=0.3\textwidth]{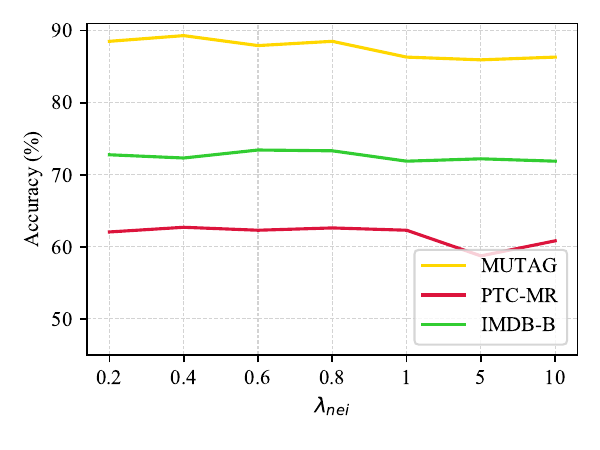}
	}
	\centering
	\caption{Experimental results of the proposed IsoC-VGAE with different values of $\lambda_{nei}$.}\label{sensitivity_nei}
\end{figure*}

\begin{figure*}[htbp]
	\centering
	\subfloat[Node Classification]{
		\includegraphics[width=0.3\textwidth]{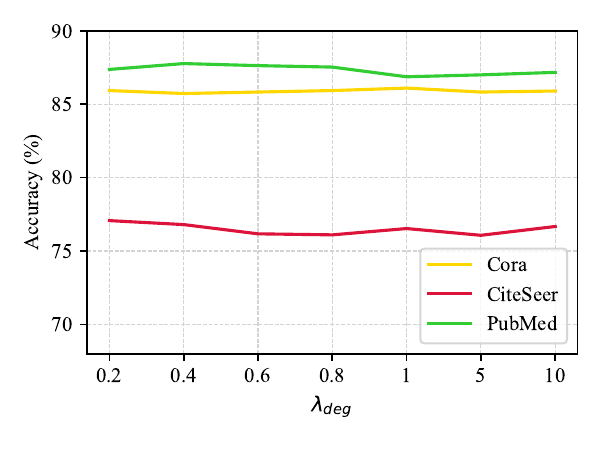}
	}
	\subfloat[Link Prediction]{
		\includegraphics[width=0.3\textwidth]{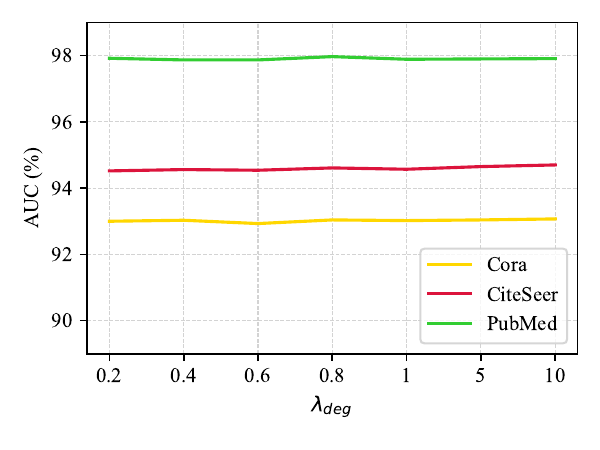}
	}
	\subfloat[Graph Classification]{
		\includegraphics[width=0.3\textwidth]{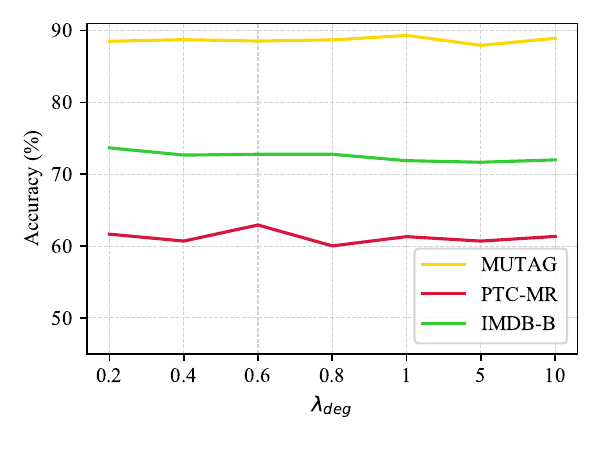}
	}
	\centering
	\caption{Experimental results of the proposed IsoC-VGAE with different values of $\lambda_{deg}$.}\label{sensitivity_deg}
\end{figure*}

In addition, we also conduct experiments of sensitivity analysis on the tuning hyperparameters of our model, including $\lambda_{nei}$ and $\lambda_{deg}$. The results are given in Fig.~\ref{sensitivity_nei} and \ref{sensitivity_deg}, which show that the performance of our model is generally robust to both of the parameters in a wide range. This is mainly because these parameters can only exert some influence on the unsupervised training process of our model, which can be almost eliminated during the fine-tuning process for training the MLPs.

\begin{table}[t]
	\centering
	\resizebox{\columnwidth}{!}{
			\begin{tabular}{lcccc}
					\toprule
					\rule{0pt}{10pt}&\textbf{Cora}&\textbf{CiteSeer}&\textbf{PubMed}&\textbf{Flickr}\\
					\midrule
					\rule{0pt}{9pt}w/o deg. \& nei.&85.4 $\pm$ 0.3&76.8 $\pm$ 0.6&87.4 $\pm$ 0.4&50.4 $\pm$ 0.1\\[1pt]
					w/o nei. &85.4 $\pm$ 0.5&76.9 $\pm$ 0.9&87.1 $\pm$ 0.4&50.5 $\pm$ 0.1\\[1pt]
					w/o deg. &85.7 $\pm$ 0.9&76.9 $\pm$ 0.8&87.6 $\pm$ 0.3&50.7 $\pm$ 0.1\\
					\midrule
					\rule{0pt}{9pt}\textbf{IsoC-VGAE}&\textbf{86.3 $\pm$ 0.4}&\textbf{77.2 $\pm$ 0.3}&\textbf{87.7 $\pm$ 0.2}&\textbf{51.2 $\pm$ 0.1}\\
					\bottomrule
				\end{tabular}
		}
	\caption{Node classification accuracy (\%) for ablation study. The best results are in bold.}\label{ablation_node}
\end{table}

\begin{table}[tb]
	\centering
	\resizebox{\columnwidth}{!}{
		\begin{tabular}{lcccc}
			\toprule
			\rule{0pt}{10pt}&\textbf{Cora}&\textbf{CiteSeer}&\textbf{PubMed}&\textbf{ogbl-collab}\\
			\midrule
			\rule{0pt}{9pt}w/o deg. \& nei.&93.1 $\pm$ 0.6&94.6 $\pm$ 0.8&97.9 $\pm$ 0.1&90.2 $\pm$ 0.1\\[1pt]
			w/o nei. &93.1 $\pm$ 0.5&94.5 $\pm$ 0.8&97.8 $\pm$ 0.2&94.7 $\pm$ 0.0\\[1pt]
			w/o deg. &93.2 $\pm$ 0.5&94.6 $\pm$ 0.9&96.6 $\pm$ 0.2&91.6 $\pm$ 0.0\\
			\midrule
			\rule{0pt}{9pt}\textbf{IsoC-VGAE}&\textbf{93.3 $\pm$ 0.3}&\textbf{95.3 $\pm$ 0.4}&\textbf{98.1 $\pm$ 0.1}&\textbf{95.2 $\pm$ 0.0}\\
			\bottomrule
		\end{tabular}
	}
        \caption{Link prediction AUC (\%) for ablation study. The best results are in bold.}\label{ablation_link}
\end{table}

\begin{table}[tb]
	\centering
	\resizebox{\columnwidth}{!}{
		\begin{tabular}{lcccc}
			\toprule
			\rule{0pt}{10pt}&\textbf{MUTAG}&\textbf{PTC-MR}&\textbf{IMDB-B}&\textbf{COLLAB}\\
			\midrule
			\rule{0pt}{9pt}w/o deg. \& nei.&85.7 $\pm$ 2.8&59.2 $\pm$ 5.1&68.9 $\pm$ 2.1&69.1 $\pm$ 0.5\\[1pt]
			w/o nei. &86.8 $\pm$ 1.6&59.1 $\pm$ 2.0&70.2 $\pm$ 1.6&68.9 $\pm$ 0.4\\[1pt]
			w/o deg. &87.9 $\pm$ 2.7&59.5 $\pm$ 2.0&69.2 $\pm$ 0.6&67.5 $\pm$ 1.9\\
			\midrule
			\rule{0pt}{9pt}\textbf{IsoC-VGAE}&\textbf{89.6 $\pm$ 2.0}&\textbf{62.5 $\pm$ 1.1}&\textbf{73.2 $\pm$ 0.9}&\textbf{70.1 $\pm$ 0.8}\\
			\bottomrule
		\end{tabular}
	}
        \caption{Graph classification accuracy (\%) for ablation study. The best results are in bold.}\label{ablation_graph}
\end{table}

\section{Conclusion}

In this paper, we propose the IsoC-VGAE for multi-level general graph representation learning. We first devise a decoding scheme to provide a theoretical guarantee for keeping the high-order isomorphic consistency of GNNs under unsupervised learning settings. Then, we propose a novel Inv-GNN decoder as an instantiation of the decoding scheme. Experimental results based on benchmark graph datasets verify that our proposed model achieves at least comparable performance to the representative graph representation learning methods on tasks at node, link and graph levels.

\bibliography{ref}
\bibliographystyle{aaai24}

\end{document}